\documentclass[letterpaper, 10 pt, conference]{ieeeconf}  

\IEEEoverridecommandlockouts                              

\let\labelindent\relax

\usepackage{mathptmx} 
\usepackage{times} 
\usepackage{amsmath} 
\usepackage{amssymb}  
\usepackage{graphicx}
\usepackage{subcaption}
\usepackage{caption}
\usepackage{booktabs}
\usepackage{multirow}
\usepackage{subcaption}
\usepackage{xcolor}
\usepackage{amsmath}
\usepackage{amssymb}
\usepackage{enumitem}
\usepackage{xcolor}
\usepackage[export]{adjustbox}
\usepackage{array}
\newcolumntype{P}[1]{>{\centering\arraybackslash}p{#1}}

\DeclareMathOperator*{\argmin}{arg\,min}
\usepackage[
  separate-uncertainty = true,
  multi-part-units = repeat
]{siunitx}
\title{\LARGE \bf
Machine Learning for Motor Learning: EEG-based Continuous Assessment of Cognitive Engagement for Adaptive Rehabilitation Robots
}

\author{Neelesh Kumar and Konstantinos P. Michmizos, Member \textit{IEEE}
\thanks{*This work is supported through Grant K12HD093427 from the National Center for Medical Rehabilitation Research, NIH/NICHD.}
\thanks{NK and KM are with the Computational Brain Lab, Department of Computer Science, Rutgers University, New Jersey, USA
        {\tt\small michmizos@cs.rutgers.edu}}%
}

\begin{document}

\maketitle
\thispagestyle{empty}
\pagestyle{empty}

\begin{abstract}
Although cognitive engagement (CE) is crucial for motor learning, it remains underutilized in rehabilitation robots, partly because its assessment currently relies on subjective and gross measurements taken intermittently. Here, we propose an end-to-end computational framework that assesses CE in real-time, using electroencephalography (EEG) signals as objective measurements. The framework consists of i) a deep convolutional neural network (CNN) that extracts task-discriminative spatiotemporal EEG to predict the level of CE for two classes- cognitively engaged vs. disengaged; and ii) a novel sliding window method that predicts continuous levels of CE in real-time. We evaluated our framework on 8 subjects using an in-house Go/No-Go experiment that adapted its gameplay parameters to induce cognitive fatigue. The proposed CNN had an average leave-one-out accuracy of 88.13\%. The CE prediction correlated well with a commonly used behavioral metric based on self-reports taken every 5 minutes ($\rho$=0.93). Our results objectify CE in real-time and pave the way for using CE as a rehabilitation parameter for tailoring robotic therapy to each patient's needs and skills.
\end{abstract}

\section{INTRODUCTION}

Rehabilitation robots have been successful in enhancing motor recovery through repetitive, goal-oriented and intensive training~\cite{krebs1998robot, krebs2000increasing, krebs1999overview}. Mounting evidence suggests that adapting the training regimes to patients' needs and skills can lead to further improvements~\cite{krebs2003rehabilitation, patton2004robot, patton2006evaluation}. Successful adaptation requires adherence to the main principles of motor learning, namely massed practice, functional relevance and cognitive engagement (CE)~\cite{damiano2006activity, lotze2003motor}, defined as the mental state that reflects the cognitive load a patient devotes towards a task. While massed practice and functional relevance have been successfully integrated in adaptive rehabilitation strategies, CE remains largely underutilized, despite having a strong relation to recovery at both the neural~\cite{lotze2003motor} and the behavioral level~\cite{krebs2003rehabilitation, patton2004robot}. 
\par
Indeed, current CE measurements have a number of limitations in driving adaptation of robotic therapy. CE has traditionally been regarded as a gross, slow-varying therapeutic variable whose range progressively includes: a) under-challenged or bored, b)  challenged and motivated, and c) over-stressed and frustrated~\cite{koenig2016human}. Recent CE measures commonly include self-reports~\cite{greene2015measuring}, i.e. feedback provided by the patients themselves. In addition to being limited by the patient's self-perception capabilities, this offline approach cannot be used for adapting therapy in real-time. Alternatively, an indirect way to assess CE is through measuring performance related to motor skills, such as the ability to initiate movements, or to reach the target accurately and in a timely manner, etc~\cite{michmizos2015robot, michmizos2012assist}; yet, these robot-derived metrics are not objective as they are often compromised by the disease itself. Physiological measurements, such as cardiovascular or skin responses, have also been used to assess CE levels~\cite{koenig2016human}; however, these signals can easily be distorted by the interference of other functions that the nervous system controls, such as homeostasis and emotions~\cite{berntson2017handbook}. Therefore, due to the lack of an objective, continuous and real-time measurement, CE remains partially explored under the scope of current adaptive rehabilitation strategies. 

\par 
Brain recordings have long been used to objectively assess the states of human behavior across the motor, cognitive and affective domains~\cite{brunovsky2003objective,sayers1974objective}. Electroencephalography (EEG) offers a non-invasive way of recording brain activity with a remarkable temporal resolution. Visual attention and cognitive workload have been shown to correlate with EEG signals~\cite{berka2007eeg, stevens2007eeg}. These studies paved the way for single-trial EEG classification of mental workload, that relied on statistical and machine learning techniques to decode the brain signals. With their main research focus being the association of EEG with simple mental components such as memory tests and problem solving tasks~\cite{liang2006classification, chaouachi2011modeling}, there is scarcity of efforts on classifying perception-action tasks that are commonly targeted in rehabilitation. In addition to not being able to generalize to new subjects, the traditional methods require computationally-intensive offline EEG pre-processing and hand-crafted features. This has hindered their incorporation in an end-to-end real-time rehabilitation system.
\par 
Recently, deep neural networks~\cite{lecun2015deep}, which have achieved tremendous success in domains such as image classification, object detection, speech synthesis, etc. have been applied to classify EEG signals for various tasks with reasonable success~\cite{li2018novel}. Such networks can automatically learn task-discriminative features and hence avoid the need for domain-specific pre-processing and hand-crafted features. In spite of having a large number of learnable parameters, these networks typically generalize very well~\cite{zhang2016understanding}. This made us wonder whether, and to what extent, a deep network would become suitable for driving a real-time rehabilitation system that adapts based on one's ability to be cognitively engaged in the motor task. 
\par 
Here, we propose an end-to-end computational framework to objectively assess CE in real-time using EEG. We first developed a convolutional neural network (CNN) that was trained to decode EEG activity and predict whether the subject was ``cognitively engaged'' or ``disengaged''. We then introduced a novel sliding window method that mapped the binary classification to a continuous level of CE in real-time. We evaluated our approach on a dataset acquired from an in-house experiment where 8 subjects performed a Go/No-Go paradigm, aimed to induce cognitive fatigue and consequent loss of engagement through adaptation of its task components. The CNN achieved average test accuracy of 88.13\% when evaluated using leave-one-out technique, and the continuous  assessment of  CE  using  the  sliding  window correlated well with a commonly used self-reported behavioral metric taken periodically (average $\rho$ = 0.93).
\begin{figure}[t]
\vspace{5.2pt}
    \centering
    \includegraphics[width=7cm]{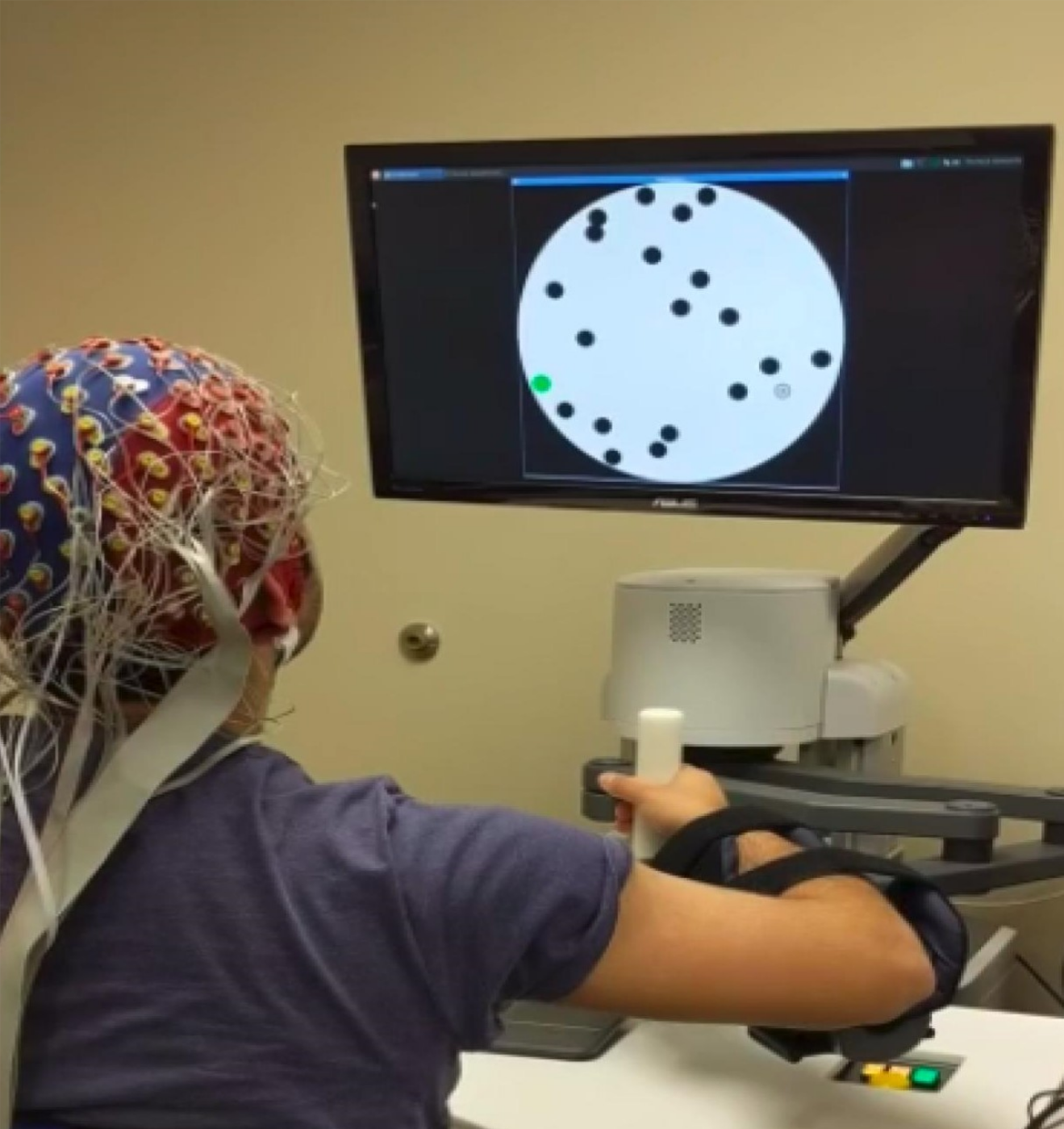}
    \caption{Experimental Setup: Subjects were asked to execute an adaptive Go/No-Go paradigm in their interaction with the Bionik InMotion Arm rehabilitation robot. EEG data were acquired as subjects performed the task.}
    \label{fig:exp}
\end{figure}
\section{Methods}
\subsection{Subjects and Experimental Apparatus}
Eight healthy volunteer subjects (age= 23 $\pm$ 2, all right handed) participated in this study that was approved by the local Institutional Review Board (IRB). The EEG data were recorded using a 128-channel Biosemi ActiveOne EEG system with a sampling frequency of 1024 Hz. The motor task was performed on the InMotion Arm Robot (Bionik Laboratories Corp.) (Figure \ref{fig:exp}). Subjects were seated at an appropriate distance from the screen so that they could perform the task comfortably without moving their torso. 

\begin{figure}[t]
\vspace{5.2pt}
    \centering
    \includegraphics[width=5cm]{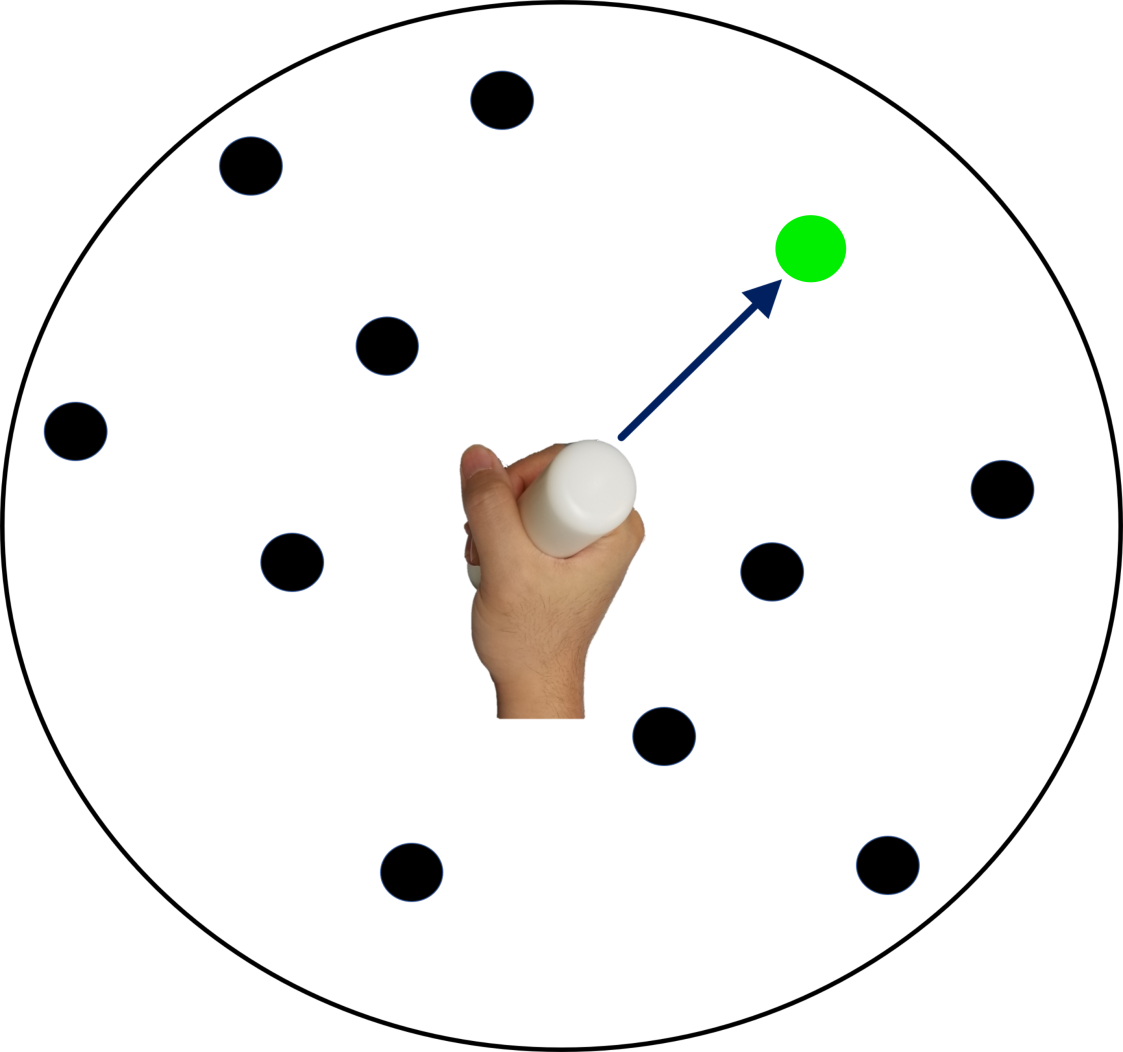}
    \caption{Our Go/No Go paradigm to induce decreasing levels of CE over time. The task consisted of a cognitive and a motor component: i) Cognitive Component: A finite number of black dots were presented on screen. After 1s, a subset of dots (2 to 5 dots randomly selected) called target dots turned blue for 2s and then returned to black. After 1s, all dots started moving for 10s. ii) Motor Component: When tracking ended, subjects responded to whether a probe dot (a random dot turned green) was one of the target dots or not; They did so by moving the robot arm towards the green dot, if the probe dot was a target dot, or not moving at all, otherwise.}
    \label{fig:game}
\end{figure}
\begin{figure}[t]
\vspace{5.2pt}
    \centering
    \includegraphics[]{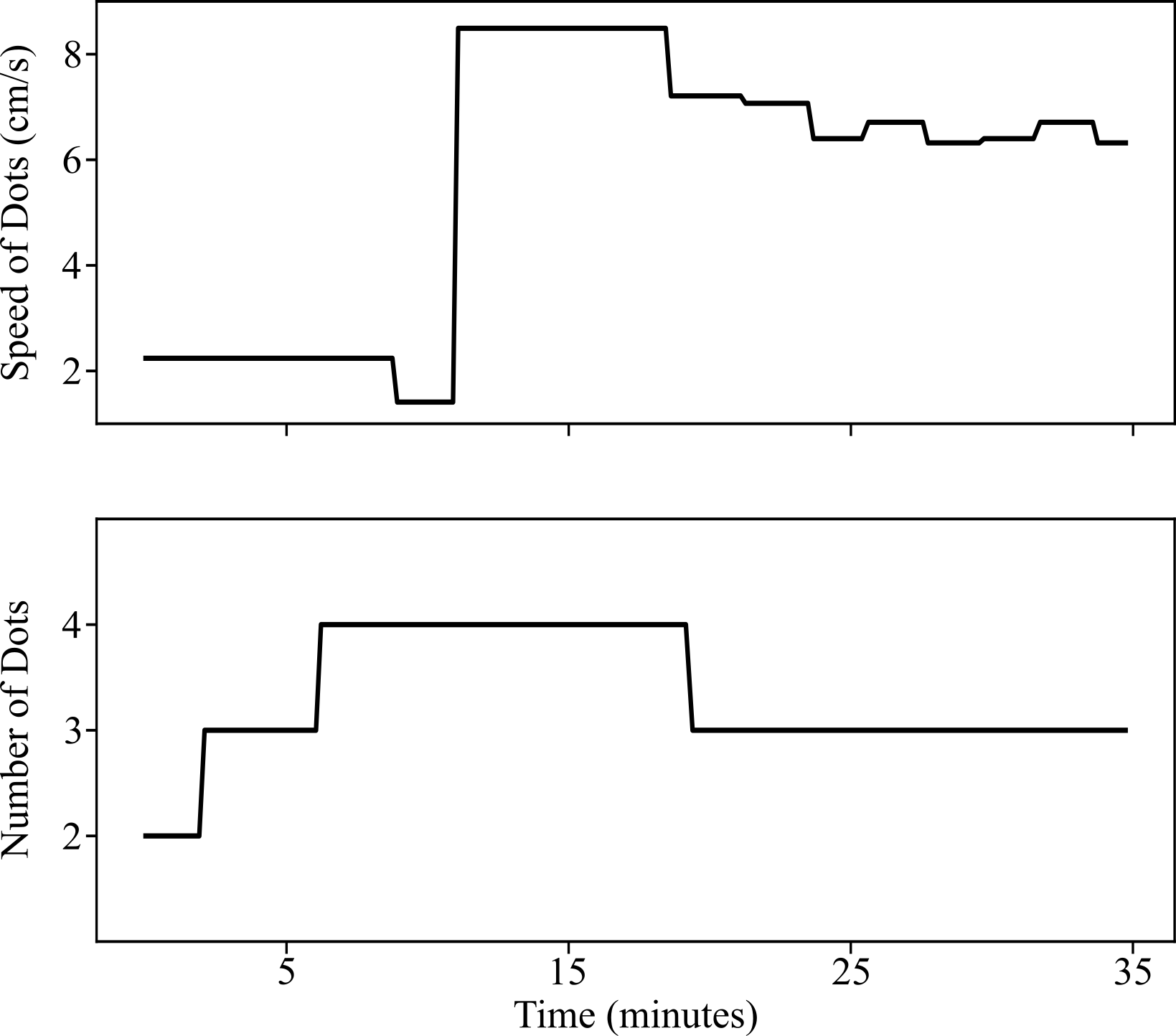}
    \caption{ Example adaptation of task parameters for one subject. The speed of the dots and the number of tracking dots were changed based on performance measures- correctness of the responses and reaction time, respectively.}
    \label{fig:behavior}
\end{figure}

\subsection{Experiment}

\begin{figure*}[t]
\vspace{5.2pt}
    \includegraphics[width=\textwidth]{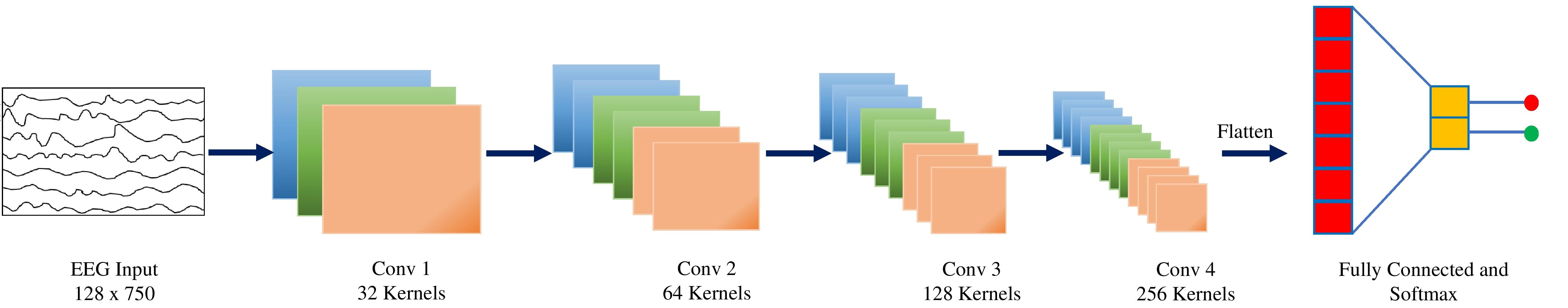}
    \caption{The 5 layer CNN Architecture. The inputs to the CNN were EEG data of dimension 128 x 600 (3s recordings from 128 channels sampled at 200Hz). The filter size for convolution was 3 x 5 for all the layers. Batch normalization and Maxpooling with filter size 2 x 3 was applied to the output of each convolutional layer. The output of the convolution layer was flattened and the resulting feature vector was passed through a fully connected layer. Finally, softmax was applied to convert the result of fully connected layer to class probabilities.}
    \label{fig:architecture}
\end{figure*}
Building on~\cite{kumar2017fatigue}, we developed an adaptive Go/No-Go task and asked the subjects to perform it to their best ability. The experimental paradigm is described in Figure \ref{fig:game}. Each subject performed the task for around 35 minutes, with the total number of trials being 170 per subject. Following a common practice~\cite{faber2012mental}, every 5 minutes, we asked the subjects to report their level of resistance against performing the task (with 1 and 9 indicating no and maximum resistance respectively). EEG data were acquired simultaneously. Our paradigm represents a typical robot-assisted goal-directed task, in which a cognitive task (e.g., motor planning) is followed by a motor task (here, a discrete movement.)

The Go/No-Go task adapted the cognitive components- the total number of dots, number of blue targets and tracking speed based on the subject's motor performance, namely correctness of responses and reaction time respectively (Figure \ref{fig:behavior}). In other words, the task difficulty (defined as the number and the speed of the dots) increased if the subjects' performance improved over time (within 11 trials), and vice-versa. By adapting to each subject's abilities, the game aimed to keep the subject maximally engaged, which resulted in decreasing levels of CE due to the inevitable cognitive fatigue induced during the 35-minutes long experiment. 
\subsection{Notations}
For each subject $i$, we created a dataset $D^i = \{(X_i^1, y_i^1), (X_i^2, y_i^2),.....,(X_i^{n_i}, y_i^{n_i})\}$, where $n_i$ denotes the number of trials recorded for that subject. For every trial $j$, $X^j \in \mathbb{R}^{128 \times T}$ is a 2 dimensional matrix which was created by stacking recordings from all the channels. $T$ is the duration of recording of each trial. The labels $y^j$ of the trial $j$ contained a value from $\{0,1\}$ corresponding to the two classes (engaged vs. disengaged). The task of the CNN was to assign class labels $y^j$ to the unseen trials.

\subsection{EEG Minimal Preprocessing and Labeling}
Minimal data preprocessing was done to get rid of the contamination in the EEG data, as described below. The minimal preprocessing on the one hand ensured the CNN would learn the representations by itself and, on the other hand,  made it possible for the CNN to be compatible with a real-time system. A bandpass filter of 0.1Hz - 40Hz was applied to remove the low and high frequency artifacts and drifts. Independent component analysis (ICA) was used to remove occular artifacts from the data. We segmented the data into trials of equal time length (3 seconds each). We then normalized the segmented trials using z-score normalization and downsampled all the trials to 200Hz to reduce the computational load. For the purpose of training the CNN, we assigned the labels to the trials in the following way: all the trials in the time window 0.5 min - 4 min were assigned the label ``cognitively engaged''; and all trials in the time window 30 min - 35 min were assigned the label ``cognitively disengaged.'' The CE dataset contained 170 data points (70 engaged and 100 disengaged) per subject. 

\subsection{The Convolutional Neural Network}
The architecture of the 5-layer CNN that we developed is shown in Figure \ref{fig:architecture}. The CNN performed convolution in spatial and temporal space. The inputs to the CNN were preprocessed EEG trials. The first 4 layers in the network were convolutional layers with the kernel size 3 x 5 and stride of 1 x 1. We applied batch normalization to the output of each convolutional layer to normalize their outputs to zero mean and unit variance, and then passed them through ReLU non-linearities. Batch normalization has been shown to provide regularization and therefore helps in avoiding overfitting \cite{ioffe2015batch}. We also applied max pooling (of window 2 x 3 with stride 2 x 3) at the end of every layer, to downsample the outputs and reduce computational load. Maxpooling has also the desirable property of translational invariance which can lead to better generalization across subjects. The last layer was a fully connected layer with softmax that took in the flattened feature vector produced by the last convolutional layer and converted it to class probabilities. The choice of the CNN hyper-parameters, i.e. the number of layers, kernel size, etc. were limited by the training data size and the input dimension, and were found using a grid search over the allowable hyper-parameters space. We performed a grid search over the allowable parameters space and selected the best ones.

\subsection{Network Training}
The CNN computed a mapping from the EEG trial to the labels, $f(X^j, \theta):\mathbb{R}^{128 \times   T}\to\mathbb\{0,1\}$ where $\theta$ are the trainable parameters of the network. The CNN was trained to minimize the average loss over all training examples:
\begin{equation}
    \hat{\theta} = \argmin \frac{1}{N} \Sigma_{i=1}^N l(X^i, y^i;\theta)
\end{equation}
where $N$ denotes the number of training examples and $l$ is the loss function, which in our case was the binary cross entropy loss function. Batch size was set to 128 and Adam (a variant of stochastic gradient descent) was used for optimization. We added dropout layer after the last convolutional layer for additional regularization. 

\subsection{Network Validation}
We evaluated our CNN in two ways: i) Leave-one-out: Data from all but one subject were used for training. Evaluation was done on the data from the left out subject. This allowed us to evaluate the ability of the CNN to generalize to new subjects, that were not included in training; ii) Subject-specific training: Data from a single subject were split randomly into training and test in the ratio 4:1. Validation was done on the test data. This tested the ability of the CNN to predict CE when trained on individual subjects.
 
\subsection{Sliding Temporal Window for Real-Time Assessment of CE}
We also propose a temporal sliding window method to measure gradual change in the level of CE. To do so, we first trained the binary classifier using the procedure outlined above. We then tested the CNN on EEG trials in a sliding window of 7 minutes with a 50 \% (3.5 minutes) overlap. The window was slid through the duration of the experiment. For each window, we computed the percentage of trials classified as 'engaged' by the trained binary classifier. This acted as a proxy for the continuous level of CE corresponding to that window.  

\section{Results}
\subsection{CNN Accuracies}
\begin{figure}[h]
\centering
    \includegraphics[]{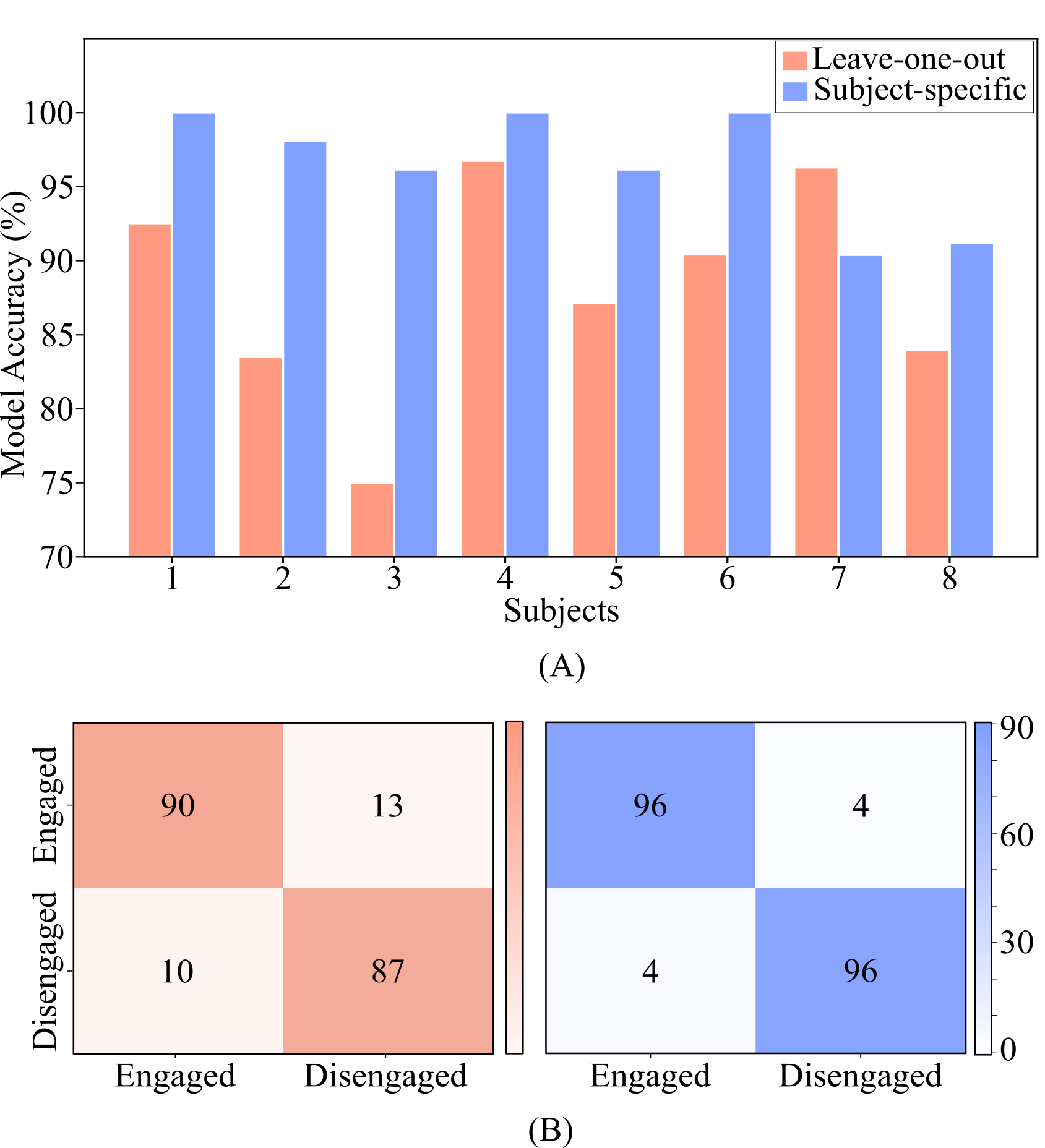}
    \caption{\textbf{A}: CNN accuracies when evaluated using leave-one-out and subject-specific training. \textbf{B}: Left- Averaged confusion matrix for leave-one-out evaluation; Right- Averaged confusion matrix for subject-specific CNNs. Numbers are in percentage. }
    \label{fig:accuracy}
\end{figure}

The results for 2-way classification of CE are shown in Figure \ref{fig:accuracy}A for both types of evaluation. When evaluated using leave-one-out, the average accuracy was $88.19\% \pm 6.86\%$. In this case, the best performance was achieved for subject 4 with an accuracy of $96.7\%$, while the lowest accuracy was obtained for subject 3 ($75\%$). Some variance is expected given the EEG signal variability across subjects. On the other hand, when evaluated using subject-specific CNN, we obtained an average accuracy of $96.49\% \pm 3.62\%$. It can be argued that if the CNN observes samples from all subjects during training, the performance is typically better than when evaluated using the leave-one-out technique, which aligns with our results. In general, we obtained, for all the subjects, accuracies that are high enough for practical usage. This is further elaborated by the confusion matrices for both evaluations (Figure \ref{fig:accuracy}B). The F1 scores (harmonic mean of precision and recall) for the two evaluations were 0.88 and 0.96, respectively. It can be observed from the matrices that the CNN predicts both classes equally well. 
\subsection{Tracking continuous loss of CE in real-time}
\begin{figure}[t]
\centering
\vspace{5.2pt}
  \includegraphics[]{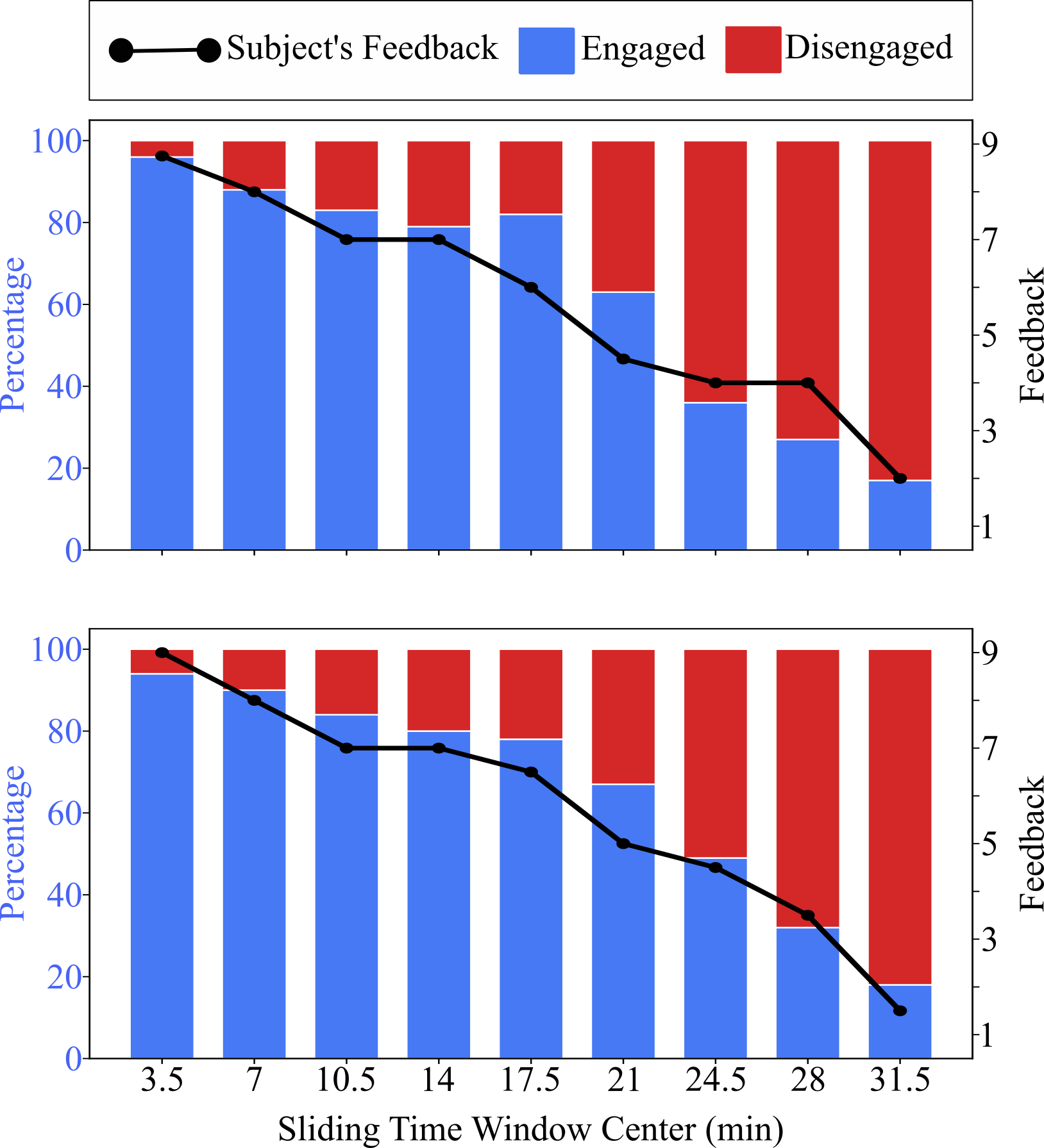}
  \caption{Continuous measurement of CE levels for two example subjects. Blue bars show the percentage of EEG trials classified as ``engaged'' in a time window. Red bars correspond to `'disengaged'' trials. Black dots indicate the subject's feedback for a given time window.}
  \label{fig:sw}
\end{figure}

We used the percentage of trials classified as ``engaged'' in a time window by the trained classifier as our metric for assessing the continuous levels of CE. Since our paradigm was designed to induce cognitive fatigue, this metric was expected to decrease with time. In Figure \ref{fig:sw}, we show the ability of our CNN to predict this gradual loss in CE. This was the case for all tested subjects. We observed that the percentage of points classified as ``cognitively engaged'' by the trained binary classifier decreased as the time window slides farther in time. Interestingly enough, the output of the sliding window correlated strongly with the self-reported feedback by the subjects indicating their level of resistance against the task, as detailed next.

\subsection{Correlation with subjects' feedback}
Every five minutes, we asked the subjects to report their resistance against continuing the task on a scale of 1 to 9. We then correlated the CE level predicted by our sliding window method with this behavioral metric by computing the Pearson's correlation coefficient. The predicted CE correlated strongly with the periodic feedback given by all subjects, with the mean $\rho$ being $0.93 \pm 0.05$.

\begin{table}[h]
\normalsize
  \begin{center}
    \caption{Pearson's Correlation Coefficient between Predicted and Self-Reported Levels of CE.}
    \label{tab:correlation}
     \begin{tabular}{|P{0.4\linewidth}|P{0.4\linewidth}|}
      \hline 
      \textbf{Subject} & \textbf{Correlation}\\
      \hline 
       1 & 0.97 \\
      2 & 0.97   \\
      3 & 0.98  \\
      4 & 0.81   \\
      5 & 0.94   \\
      6 & 0.96  \\
      7 & 0.95   \\
      8 & 0.88  \\
      \hline
      \textbf{Mean} & \SI{0.93 \pm 0.05}{}  \\
      \hline 
    \end{tabular}
  \end{center}
\end{table}
\section{Discussion}
In this paper, we presented a computational framework that couples EEG-informed continuous assessment of CE with rehabilitation robots. To demonstrate our method's applicability, we validated the CNN's ability to predict CE when a) both the training and the validation EEG data belonged to the same subject (subject-specific), and b) data from the test subject were not used for training the CNN (leave-one-out). The high accuracies reported in both the cases signify our ongoing effort to tailor robotic therapy based on the real-time assessment of CE. 
\par
In that sense, the knowledge of CE aims to complement the robot-derived quantitative metrics of motor performance~\cite{bosecker2010kinematic, krebs2014robotic}, already corroborating that successful motor recovery not only requires repetitive movements~\cite{krebs2016beyond}, but ``repetition without repetition''~\cite{bernstein1966co} signifying ``active participation'' during therapy~\cite{krebs2003rehabilitation, patton2004robot}. Indeed, a patient highly engaged in a challenging rehabilitation environment is likely to participate actively in the motor task, which is known to enhance motor learning and thus improve clinical outcomes~\cite{lotze2003motor}. Given that both upper and lower extremity discrete movements are likely controlled by the same sensorimotor control strategies~\cite{michmizos2014comparative}, the real-time knowledge of CE can be applicable to a wide range of rehabilitation tasks. 
\par 
Our validation method used self-reports as ``data points'' in measure (level of CE) and time, but went beyond simply correlating the predicted and the self-reported levels of CE. A main disadvantage of self-reports (questionnaires) associated with changes in mental states~\cite{boucsein2008methods} is that they provide information at discrete points in time, typically after the completion of the training, and cannot be used in real-time. Further, many neurological patients, particularly children, often suffer from cognitive deficits that limit their self-perception capabilities and cannot objectively report their levels of CE. Our sliding window method consistently predicted the decreasing levels of CE over time. Its strong correlation with the self-reported levels of CE across all subjects is an additional evidence that it can be used to assess CE at any time instant during therapy. 
\par
Adding to the mounting evidence that deep networks can predict cognitive functions~\cite{craik2019deep}, our work showed how CE can be measured using EEG as an objective signal. We also showed the ability of deep networks to extract task-relevant features from noisy EEG signals with minimal pre-processing of the data. The leave-one-out evaluation results suggest that our CNN generalizes well across all the subjects. This is important as the system can be used for new patients with minimal or no need to retrain the CNN.
\par 
An explicit measurement of CE levels during therapy will help us devise new adaptive rehabilitation strategies. For example, we have previously shown how cognitive components of a serious game can be adapted to promote patient engagement during sensorimotor therapy~\cite{kommalapati2016virtual}. In addition to opening up the possibilities of adapting gamification to the CE level, the proposed framework can also augment the assist-as-needed adaptive controllers that we have previously proposed ~\cite{michmizos2012assist,michmizos2015robot,michmizos2017pediatric}, by separating the cognitive component from the motor components of the rehabilitation tasks. In other words, the real-time knowledge of CE based on EEG signals is unbiased by the motor state of the disease and can allow further exploration of the conditions under which patients exhibit the largest capacity to learn. Our framework could also help in identifying a ``therapeutic window'', i.e. task conditions that maximize CE. Under this therapeutic window, patients can potentially exhibit their largest capacity in motor learning, which could result in more efficient rehabilitation strategies. 
\par
Overall, our results pave the way for using CE as a rehabilitation parameter for tailoring robotic therapy to each patient's needs and skills. A fast-responsive CE scale can also offer new insights on the long-proven association of CE to motor recovery and allow for further personalizing the evidence-based rehabilitation robots and clarifying the underlying motor neuroscience. While thorough studies are needed, a real-time CE assessment can help in answering a number of open questions: What game parameters should we adapt to find the right responders or get the maximum improvement? What ``dosage'' should we provide to each patient? What kind of feedback should we give? Is CE related to implicit \cite{michmizos2015robot} or explicit motor learning? Would a type of social interaction during a gameplay be beneficial? Answering these and other questions would help us study the origin of motor learning and offer novel and testable hypotheses that combine CE with neurorehabilitation robotics.

\section{Conclusion}
This work aims to complement the current approaches where robots are simultaneously used as therapeutic and performance-evaluation devices, by introducing an objective assessment of CE that can be used to adapt therapy. The EEG-based CE measurement, which is unbiased by the motor state of the disease, is expected to allow exploration of the conditions under which patients exhibit their largest capacity to learn. Its further integration with the commonly used motor behavioral metrics could help us step away from evidence-based and move towards science-based rehabilitation, increasing our insights on motor recovery and facilitating our efforts to harness and nurture brain plasticity.
\bibliographystyle{IEEEtran}
\bibliography{ref}
\end{document}